\documentclass[11pt,a4paper]{article}
\usepackage[hyperref]{naaclhlt2018}
\usepackage{times}
\usepackage{latexsym}
\usepackage{color}
\usepackage{graphicx}
\usepackage{amsmath}
\usepackage{multirow}
\usepackage[noend]{algcompatible}
\usepackage{algorithm}
\usepackage{url}

\aclfinalcopy 
\def\aclpaperid{523} 

\title{Guiding Neural Machine Translation with Retrieved Translation Pieces}

\author{Jingyi Zhang$^{1,2}$, Masao Utiyama$^1$, Eiichro Sumita$^1$\\ \bf{Graham Neubig}$^{3,2}$, Satoshi Nakamura$^2$\\
$^1$National Institute of Information and Communications Technology,  Japan\\
$^2$Graduate School of Information Science,
Nara Institute of Science and Technology,  Japan\\
$^3$Language Technologies Institute, Carnegie Mellon University, USA \\
  {\tt   jingyizhang/mutiyama/eiichiro.sumita@nict.go.jp }\\
  {\tt  gneubig@cs.cmu.edu, s-nakamura@is.naist.jp}
}

\date{}

\begin{document}
\maketitle
\begin{abstract}
One of the difficulties of neural machine translation (NMT) is the recall and appropriate translation of low-frequency words or phrases.
In this paper, we propose a simple, fast, and effective method for recalling previously seen translation examples and incorporating them into the NMT decoding process.
Specifically, for an input sentence, we use a search engine to retrieve sentence pairs whose source sides are similar with the input sentence, and then collect   $n$-grams that are both in the retrieved target sentences and aligned with words that match in the source sentences, which we call ``translation pieces''.
We compute pseudo-probabilities for each retrieved sentence based on similarities between the input sentence and the retrieved source sentences, and use these to weight the retrieved translation pieces.
Finally, an existing NMT model is used to translate the input sentence, with an additional bonus given to outputs that contain the collected translation pieces.
We show our method improves NMT translation results up to 6 BLEU points on three narrow domain translation tasks where repetitiveness of the target sentences is particularly salient.
It also causes little increase in the translation time, and compares favorably to another alternative retrieval-based method with respect to accuracy, speed, and simplicity of implementation.
\end{abstract}

\section{Introduction}

Neural machine translation (NMT) \cite{bahdanau2014neural,sennrich-haddow-birch:2016:WMT,wang-EtAl:2017:WMT} is now the state-of-the-art in machine translation, due to its ability to be trained end-to-end on large parallel corpora and capture complex parameterized functions that generalize across a variety of syntactic and semantic phenomena.
However, it has also been noted that compared to alternatives such as phrase-based translation \cite{koehn2003statistical}, NMT has trouble with low-frequency words or phrases \cite{arthur-neubig-nakamura:2016:EMNLP2016,kaiser2017learning}, and also generalizing across domains \cite{koehn-knowles:2017:NMT}.
A number of methods have been proposed to ameliorate these problems, including methods that incorporate symbolic knowledge such as discrete translation lexicons \cite{arthur-neubig-nakamura:2016:EMNLP2016,he2016improved,chatterjee-EtAl:2017:WMT1} and phrase tables \cite{zhang-EtAl:2017:Long2,tang2016neural,dahlmann-EtAl:2017:EMNLP2017}, adjust model structures to be more conducive to generalization \cite{nguyen2017improving}, or incorporate additional information about domain \cite{wang-EtAl:2017:Short3} or topic \cite{zhang-EtAl:2016:COLING3} in translation models.

 \begin{figure*}[t]
   \center
   \includegraphics[width=0.99\textwidth]{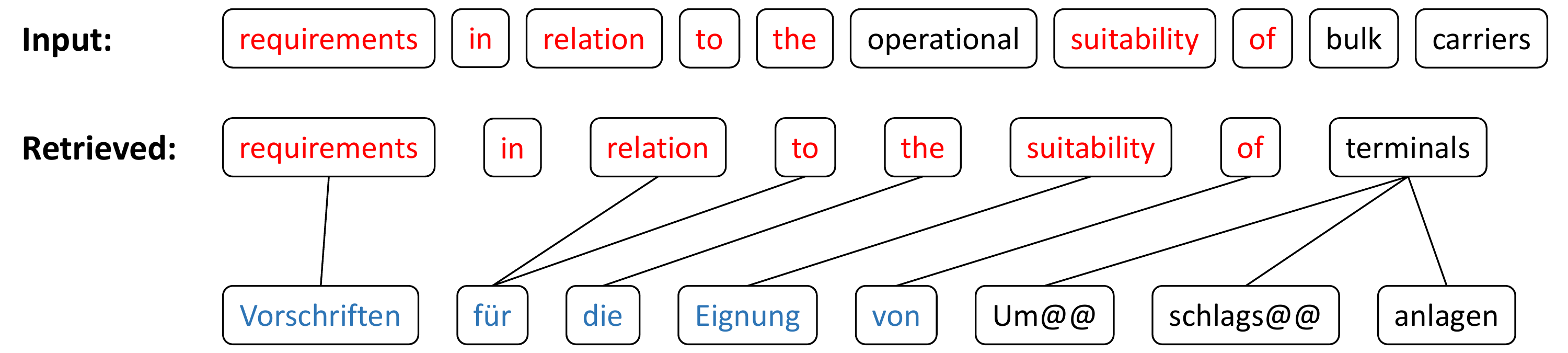}
   \caption{A word-aligned sentence pair retrieved for an input sentence. Red words are unedited words obtained by computing the edit distance between the input sentence and the retrieved source sentence. The blue part of the retrieved target sentence is collected as translation pieces for the input sentence. The target word ``Umschlagsanlagen" is split into ``Um@@", ``schlags@@" and ``anlagen" by byte pair encoding.}
   \label{f1}
  \end{figure*}

In particular, one paradigm of interest is recent work that augments NMT using \emph{retrieval}-based models, retrieving sentence pairs from the training corpus that are most similar to the sentence that we want to translate, and then using these to bias the NMT model.\footnote{Note that there are existing retrieval-based methods for phrase-based and hierarchical phrase-based translation \cite{lopez:2007:EMNLP-CoNLL2007,germann2015sampling}. However, these methods do not improve translation quality but rather aim to improve the efficiency of the translation models.}
These methods -- reminiscent of translation memory \cite{utiyama2011searching} or example-based translation \cite{nagao1984framework,grefenstette1999world} -- are effective because they augment the parametric NMT model with a non-parametric translation memory that allows for increased capacity to measure features of the target technical terms or domain-specific words.
Currently there are two main approaches to doing so.
\newcite{li2016one} and \newcite{farajian-EtAl:2017:WMT} use the retrieved sentence pairs to fine tune the parameters of the NMT model which is pre-trained on the whole training corpus.
\newcite{gu2017search} uses the retrieved sentence pairs as additional inputs to the NMT model to help NMT in translating the input sentence.
While both of these paradigms have been proven  effective, they both add significant complexity and computational/memory cost to the decoding process, and also to the training procedure.
The first requires the running of several training iterations and rolling back of the model, which is costly at test time, and the second requires entirely changing the model structure which requires training the model separately, and also increases test-time computational cost by adding additional encoders.

In this paper, we propose a simple and efficient model for using retrieved sentence pairs to guide an existing NMT model at test time.
Specifically, the model collects $n$-grams occurring in the retrieved target sentences that also match words that overlap between the input and retrieved source sentences, which we will refer to as ``translation pieces'' (e.g., in Figure~\ref{f1}, the blue part of the retrieved target sentence is collected as translation pieces for the input sentence).
The method then calculates a pseudo-probability score for each of the retrieved example sentence pairs and weights the translation pieces according to this value.
Finally, we up-weight NMT outputs that contain the collected translation pieces.
Unlike the previous methods, this requires no change of the underlying NMT model and no updating of the NMT parameters, making it both simple and efficient to apply at test time.

We show our method improved NMT translation results up to 6 BLEU points on three translation tasks and caused little increase in the translation time.
Further, we find that accuracies are comparable with the model of \newcite{gu2017search}, despite being significantly simpler to implement and faster at test time.

\section{Attentional NMT}
Our baseline NMT model is similar to the attentional model of \newcite{bahdanau2014neural}, which includes an encoder, a decoder and an attention (alignment) model.
Given a source sentence $X = \left\{ {{x_1},...,{x_L}} \right\}$, the encoder learns an annotation ${h_i} = \left[ {{{\vec h}_i};{{\mathord{\buildrel{\lower3pt\hbox{$\scriptscriptstyle\leftarrow$}} 
\over h} }_i}} \right]$ for $x_i$ using a bi-directional recurrent neural network. 

The decoder generates the target translation from left to right. The probability of generating next word $y_t$ is,\footnote{$g$, $f$ and $a$ in Equation~\ref{e1}, \ref{e2} and \ref{e4} are nonlinear, potentially multi-layered, functions.}
\begin{equation}
P_{NMT}\left( {{y_t}|y_1^{t - 1},X} \right) = softmax\left( {g\left( {{y_{t - 1}},{z_t},{c_t}} \right)} \right)
\label{e1}
\end{equation} 
where $z_t$ is a decoding state for time step $t$, computed by, 
\begin{equation} 
{z_t} = f\left( {{z_{t - 1}},{y_{t - 1}},{c_t}} \right)
\label{e2}
\end{equation}
 $c_t$ is a source representation for time $t$, calculated as,
\begin{equation} 
{c_t} = \sum\limits_{i = 1}^L {{\alpha _{t,i}} \cdot {h_i}} 
\label{e3}
\end{equation} 
where ${\alpha _{t,i}}$ scores how well the inputs around position $i$ and the output at position
$t$ match, computed as,
\begin{equation} 
{\alpha _{t,i}} = \frac{{\exp \left( {a\left( {{z_{t - 1}},{h_i}} \right)} \right)}}{{\sum\limits_{j = 1}^L {\exp \left( {a\left( {{z_{t - 1}},{h_j}} \right)} \right)} }}
\label{e4}
\end{equation}
The standard decoding algorithm for NMT is beam search. That is, at each time step $t$, we keep $n$-best hypotheses. The probability of a complete hypothesis is computed as,
\begin{equation}
\log {P_{NMT}}\left( {Y|X} \right) = \sum\limits_{t = 1}^{\left| Y \right|} {\log {P_{NMT}}\left( {{y_t}|y_1^{t - 1},X} \right)} 
\label{e5}
\end{equation}
Finally, the translation score is normalized by sentence length to avoid too short outputs.
\begin{equation}
\log {S_{NMT}}\left( {Y|X} \right) = \frac{{\log {P_{NMT}}\left( {Y|X} \right)}}{{\left| Y \right|}}
\label{e6}
\end{equation}

\section{Guiding NMT with Translation Pieces}
\label{our}
This section describes our approach, which mainly consists of two parts: 
\begin{enumerate}
\item retrieving candidate translation pieces from a parallel corpus for the new source sentence that we want to translate, and then
\item using the collected translation pieces to guide an existing NMT model while translating this new sentence.
\end{enumerate}
At training time, we first prepare  the parallel corpus that will form our database used in the retrieval of the translation pieces.
Conceivably, it could be possible to use a different corpus for translation piece retrieval and NMT training, for example when using a separate corpus for domain adaptation, but for simplicity in this work we use the same corpus that was used in NMT training.
As pre-processing, we use an off-the-shelf word aligner to learn word alignments for the parallel training corpus.

\subsection{Retrieving Translation Pieces}

At test time we are given an input sentence $X$.
For this $X$, we first use the off-the-shelf search engine Lucene to search the word-aligned parallel training corpus and retrieve $M$ source sentences $\left\{ {{X^m}:1 \le m \le M} \right\}$ that are similar to $X$.
$Y^m$ indicates the target sentence that corresponds to source sentence $X^m$ and $\mathcal A^m$ is word alignments between $X^m$ and $Y^m$. 

For each retrieved source sentence $X^m$, we compute its edit distance with $X$ as ${d\left( {X,{X^m}} \right)}$ using dynamic programming.
We record the unedited words in $X^m$ as $\mathcal W^m$, and also note the words in the target sentence $Y^m$ that correspond to source words in $\mathcal W^m$, which we can presume are words that will be more likely to appear in the translated sentence for $X$.
According to Algorithm~\ref{a1}, we collect  $n$-grams (up to $4$-grams) from the retrieved target sentence $Y^m$ as possible translation pieces $ G_X^m$ for $X$, using word-level alignments to select $n$-grams that are related to $X$  and discard $n$-grams that are not related to $X$. 
The final translation pieces $G_X$ collected for $X$ are computed as,\footnote{Note that the extracted translation pieces are target phrases, but the target words contained in one extracted translation piece may be aligned to discontiguous source words, which is different from how phrase-based translation extracts phrase-based translation rules.}
\begin{equation}
 G{_X} = \bigcup\limits_{m = 1}^M { G_X^m} 
\label{e7}
\end{equation}

 Table~\ref{t-our} shows a few $n$-gram examples contained in the retrieved target sentence in Figure~\ref{f1} and whether they are included in $G_X^m$ or not. Because the retrieved source sentence in Figure~\ref{f1} is highly similar with the input sentence, the translation pieces collected from its target side are highly likely to be correct translation pieces of the input sentence.
 However, when a retrieved source sentence is not very similar with the input sentence (e.g. only one or two words match), the translation pieces collected from its target side will be less likely to be correct translation pieces for the input sentence.
 
We compute a score for each  $u \in G{_X}$ to measure how likely it is a correct translation piece for $X$ based on sentence similarity between the retrieved source sentences and the input sentence as following,
\begin{equation}
\begin{gathered}
S\left( {u,X,\bigcup\limits_{m = 1}^M {\left\{ {\left( {{X^m},G_X^m} \right)} \right\}} } \right) \\= \mathop {\max }\limits_{1 \le m \le M \wedge  u \in G_X^m} simi\left( {X,{X^m}} \right)
\end{gathered}
\label{e8}
\end{equation}
where $simi\left( {X,{X^m}} \right)$ is the sentence similarity computed as following \cite{gu2017search},
\begin{equation}
simi\left( {X,{X^m}} \right) = 1 - \frac{{d\left( {X,{X^m}} \right)}}{{\max \left( {\left| X \right|,\left| {{X^m}} \right|} \right)}}
\label{e9}
\end{equation}

 \begin{algorithm}[t]
  \caption{Collecting Translation Pieces }
  \begin{algorithmic}
   \REQUIRE $X=x_1^{L}$, $X^m=k_1^{L'}$, $Y^m=v_1^{L''}$, $\mathcal A^m$, $\mathcal W^m$
   \ENSURE $ G_{X}^m$
     \STATE $ G_X^m = \emptyset $

       \FOR {$i=1$ to $L''$}
          \FOR {$j=i$ to $L''$}
            \IF {$j-i=4$}
                \STATE \textbf{break}
              \ENDIF 
               \IF {$\exists p$ $:$ $\left( {p,j} \right) \in {\mathcal A^m}$ $\wedge $ $p \notin {\mathcal W^m}$}
                \STATE \textbf{break}
              \ENDIF 
       \STATE add $v_i^j$ into $ G_X^m$
        
             \ENDFOR  
         \ENDFOR  
  
  \end{algorithmic}
  \label{a1}
\end{algorithm}

 \begin{table}[t]
 \centering
\begin{tabular}{l|l}
\hline
$n$-grams& $G_X^m$\\
\hline 
Vorschriften f\"ur die Eignung&Yes\\
die Eignung von&Yes\\
von Um@@ schlags@@ anlagen&No\\
Um@@ schlags@@ anlagen&No\\
\hline
\end{tabular}
\caption{Examples of the collected translation pieces.}
\label{t-our}
\end{table}

\subsection{Guiding NMT with Retrieved Translation Pieces}

\begin{figure}[t]
   \center
   \includegraphics[width=0.5\textwidth]{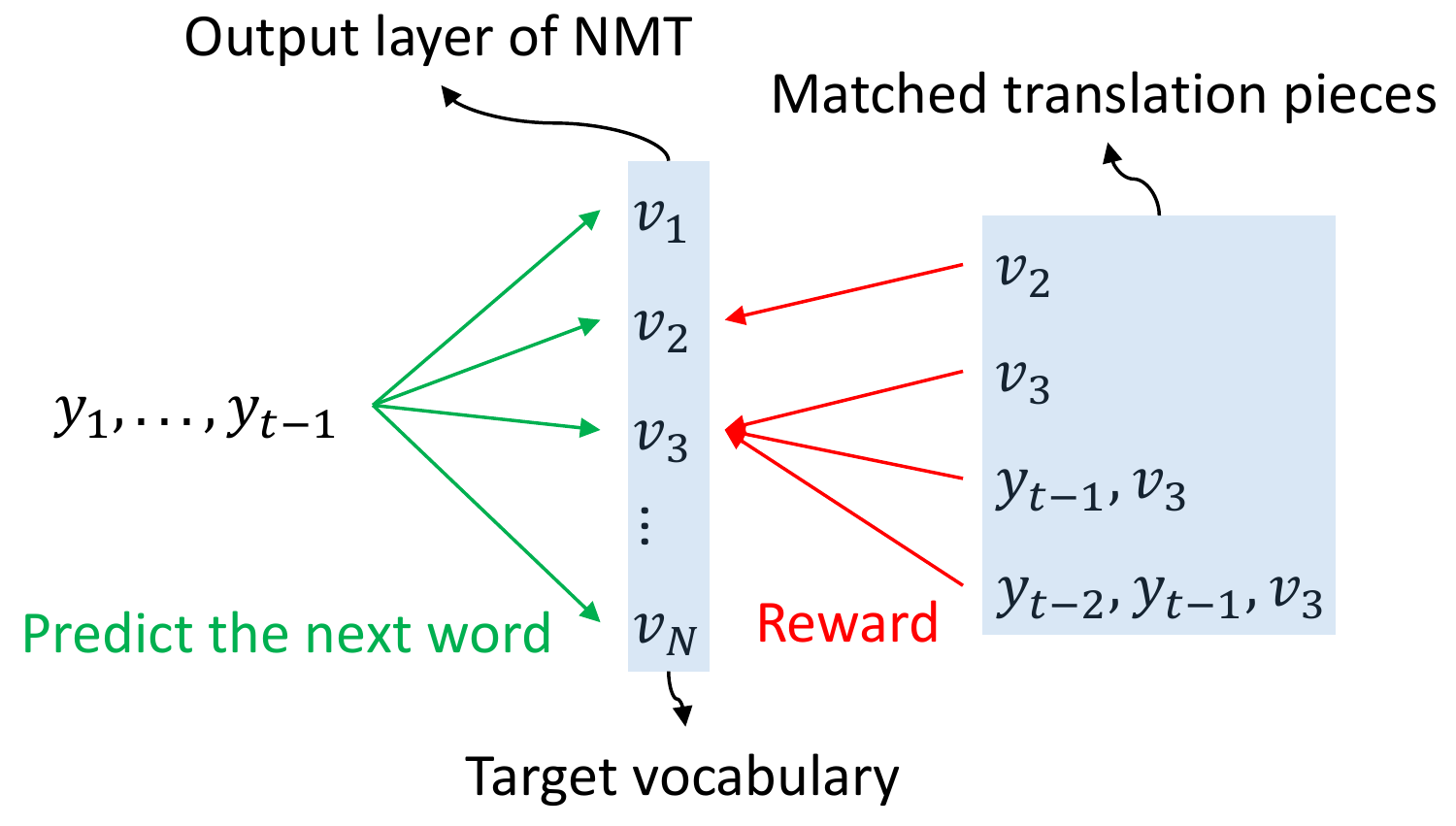}
   \caption{A simple demonstration of adding rewards for matched translation pieces into the NMT output layer.}
   \label{simple-demo}
  \end{figure}

In the next phase,  we use our NMT system to translate the input sentence. Inspired by \newcite{stahlberg-EtAl:2017:EACLshort} which rewards $n$-grams from syntactic translation lattices during NMT decoding, we add an additional reward for $n$-grams that occur in the collected translation pieces.
That is, as shown in Figure~\ref{simple-demo}, at each time step $t$, we update the  probabilities over the output vocabulary and increase the probabilities of those that result in matched $n$-grams according to
\begin{equation} 
\begin{gathered}
\log S_{NMT\_updated}\left( {{y_t}|y_1^{t - 1},X} \right) \\= \log {P_{NMT}}\left( {{y_t}|y_1^{t - 1},X} \right) + \\ \lambda \sum\limits_{n = 1}^4 {\delta \left( {y_{t - n + 1}^t,X,\bigcup\limits_{m = 1}^M {\left\{ {\left( {{X^m},G_X^m} \right)} \right\}} } \right)} ,
\end{gathered}
\label{e10}
\end{equation}
where $\lambda $ can be tuned on the development set and $\delta \left(  \cdot  \right)$ is computed as Equation~\ref{e8} if $y_{t-n+1}^{t} \in G_X$, otherwise $\delta \left(  \cdot  \right)=0$.

To implement our method, we use a dictionary $\mathcal D_X$ to store translation pieces $ G_X$ and their scores for each input sentence $X$. At each time step $t$, we update the output layer probabilities by checking $\mathcal D_X$.
However, it is inefficient to traverse all target words in the vocabulary and check whether they belong to $ G_X$ or not, because the vocabulary size is large. Instead, we only traverse target words that belong to $ G_X$ and update the corresponding output probabilities as shown in Algorithm~\ref{a2}. Here, $\mathcal L_X$ is a list that stores $1$-grams contained in $ G_X$.\footnote{Note that our method does not introduce new states during decoding, because the output layer probabilities are simply updated based on history words and the next word.}

\begin{algorithm}[t] 
  \caption{Guiding NMT by Translation Pieces }
  \begin{algorithmic}
   \REQUIRE Output layer $\log P_{NMT}\left( {{y_t}|y_1^{t - 1},X} \right) $, $\mathcal L_X$, $\mathcal D_X$
   \ENSURE Updated output layer

       \FOR {$u$ in $\mathcal L_X$}
        \STATE $\log {P_{NMT}}\left( {u|y_1^{t - 1},X} \right) + = \lambda \mathcal D_X\left( u \right)$
          \FOR {$i=1$ to $3$}
            \IF {$t-i<1$}
                \STATE \textbf{break}
                   \ENDIF 
                 \IF { $y_{t-i}^{t-1}, u$ $\notin $ $\mathcal D_X$}
                \STATE \textbf{break}
              \ENDIF 
           
                \STATE $\log {P_{NMT}}\left( {u|y_1^{t - 1},X} \right) + $$= $$ \lambda \mathcal D_X\left( y_{t-i}^{t-1},u \right)$

             \ENDFOR  
         \ENDFOR  
  \end{algorithmic}
  \label{a2}
 \end{algorithm}

\begin{table*}[t] 
\center
\begin{tabular}{ll|ll|ll|ll}
\hline
&&en-de&&en-fr&&en-es&\\
&&BLEU&METEOR&BLEU&METEOR&BLEU&METEOR\\
\hline
dev&NMT&44.08&36.69&57.26&43.51&55.76&42.53\\
&Ours&50.81&39.50&62.60&45.83&60.51&44.58\\
\hline
test&NMT&43.76&36.57&57.67&43.66&55.78&42.55\\
&Ours&50.15&39.18&63.27&46.24&60.54&44.64\\
\hline
\end{tabular}
\caption{Translation results.}
\label{t-result}
\end{table*}

As we can see, our method only  up-weights NMT outputs that match the retrieved translation pieces in the NMT output layer. In contrast, \newcite{li2016one} and \newcite{farajian-EtAl:2017:WMT} use the retrieved sentence pairs to run additional training iterations and fine tune the NMT parameters for each input sentence; \newcite{gu2017search} runs the NMT model for each retrieved sentence pair to obtain the NMT encoding and decoding information of the retrieved sentences as key-value memory to guide NMT for translating the new input sentence. Compared to their methods, our method adds little  computational/memory cost and is simple to implement.

\section{Experiment}
\subsection{Settings}
Following \newcite{gu2017search}, we use version 3.0 of the JRC-Acquis corpus for our translation experiments.
The JRC-Acquis corpus 
 contains the total body of European Union (EU) law applicable in the EU Member States. It can be used as a narrow domain to test the effectiveness of our proposed method. We did translation experiments on three directions: English-to-German (en-de), English-to-French (en-fr) and English-to-Spanish (en-es).

 We cleaned the data by removing repeated sentences and used the \texttt{train-truecaser.perl} script from Moses \cite{koehn-EtAl:2007:PosterDemo} to truecase the corpus. Then we selected 2000 sentence pairs as development and test sets, respectively. The rest was used as the training set. We removed
 sentences longer than 80 and 100 from the training and development/test sets respectively. 
 The final numbers of sentence pairs contained in the training, development and test sets are shown in Table~\ref{t-data}.\footnote{We put the datasets used in our experiments on Github
 
  https://github.com/jingyiz/Data-sampled-preprocessed} We applied \textit{byte pair encoding} \cite{sennrich-haddow-birch:2016:P16-12} and set the vocabulary size to be 20K.
\begin{table}[t]  
\center
\begin{tabular}{l|lll}
\hline
&en-de&en-fr&en-es\\
\hline
TRAIN&674K&665K&663K\\
DEV&1,636&1,733&1,662\\
TEST&1,689&1,710&1,696\\
Average Length&31&29&29\\
\hline
\end{tabular}
\caption{Data sets. The last line is the average length of English sentences.}
\label{t-data}
\end{table}
 
For translation piece collection, we use GIZA++
 \cite{och2003systematic} and the \texttt{grow-diag-final-and}
 heuristic \cite{koehn2003statistical} to obtain
 symmetric word alignments for the training set.
 
 We trained an attentional NMT model as our baseline system. The settings for NMT are shown in Table~\ref{t-para}. We also compared our method with the search engine guided NMT model (SGNMT, \newcite{gu2017search}) in Section~\ref{sgnmt}.

 \begin{table}[H] \small
 
 \center
 \begin{tabular}{l|l}
 \hline
Word embedding &512 \\
GRU dimension& 1024 \\
 Optimizer &adam\\
 Initial learning rate &0.0001\\
 Beam size& 5\\
 \hline
 
 \end{tabular}
 \caption{ NMT settings.}
 \label{t-para}
 \end{table}

 For each input sentence, we retrieved 100 sentence pairs from the training set using Lucene as our preliminary setting. We analyze the influence of the retrieval size in Section~\ref{comp-plex}. 
 The weights of translation pieces used in Equation~\ref{e10} are tuned on the development set for different language pairs, resulting in weights of 1.5 for en-de and en-fr, and a weight of 1 for en-es.

\begin{table}[t]  
\center
\begin{tabular}{ll|lll}
\hline
&&en-de&en-fr&en-es\\
\hline
dev&NMT&1.000&0.990&0.997\\
&Ours&1.005&0.991&1.001\\
\hline
test&NMT&0.995&0.990&0.990\\
&Ours&1.004&0.989&0.993\\
\hline
\end{tabular}
\caption{Ratio of translation length to reference length.}
\label{t-ratio}
\end{table}

\subsection{Results}
Table~\ref{t-result} shows the main experimental results.
We can see that our method outperformed the baseline NMT system up to 6 BLEU points.
As large BLEU gains in neural MT can also often be attributed to changes in output length, we examined the length (Table~\ref{t-ratio}) and found that it did not influence the translation length significantly.

In addition, it is of interest whether how well the retrieved sentences match the input influences the search results.
We measure the similarity between a test sentence $X$ and the training corpus $D_{train}$ by computing the sentence similarities between $X$ and the retrieved source sentences as
\begin{equation} 
simi\left( {X,D_{train}} \right) = \mathop {\max }\limits_{1 \le m \le M} simi\left( {X,{X^m}} \right).
\label{simi1n}
\end{equation}
 The similarity between the test set $D_{test}$ and the training corpus $D_{train}$ is measured as, 
\begin{equation}\small 
simi\left( {{D_{test}},{D_{train}}} \right) = \frac{{\sum\nolimits_{X \in {D_{test}}} {simi\left( {X,{D_{train}}} \right)} }}{{\left| {{D_{test}}} \right|}}
\label{siminn}
\end{equation}

Our analysis demonstrated that, expectedly, the performance of our method is highly influenced by the similarity between the test set and the training set.
We divided sentences in the test set into two parts: half has higher similarities with the training corpus (half-H) and half has lower similarities with the training corpus (half-L).
Table~\ref{t-simi} shows the similarity between the training corpus and the whole/divided test sets.
Table~\ref{t-simi-bleu} shows translation results for the whole/divided test sets.
As we can see, NMT generally achieved better BLEU scores for half-H and our method improved BLEU scores for half-H much more significantly than for half-L, which shows our method can be quite useful for narrow domains where similar sentences can be found.

\begin{table}[t]   
\center
\begin{tabular}{l|lll}
\hline
&whole& half-H& half-L\\
\hline
en-de&0.56&0.80&0.32\\
en-fr&0.57&0.81&0.33\\
en-es&0.57&0.81&0.32\\
\hline
\end{tabular}
\caption{Similarities between the training set and the whole/divided test sets.}
\label{t-simi}
\end{table}

\begin{table}[t]  
\center
\begin{tabular}{ll|lll}
\hline
&&whole&half-H&half-L\\
\hline
en-de&NMT&43.76&60.93&32.25\\
&Ours&50.15&73.26&34.28\\
\hline
en-fr&NMT&57.67&72.64&47.38\\
&Ours&63.27&82.76&49.81\\
\hline 
en-es&NMT&55.78&69.32&46.26\\
&Ours&60.54&78.37&47.93\\
\hline
\end{tabular}
\caption{Translation results (BLEU) for the whole/divided test sets.}
\label{t-simi-bleu}
\end{table}

 \begin{figure*}[t]
   \center
   \includegraphics[width=1\textwidth]{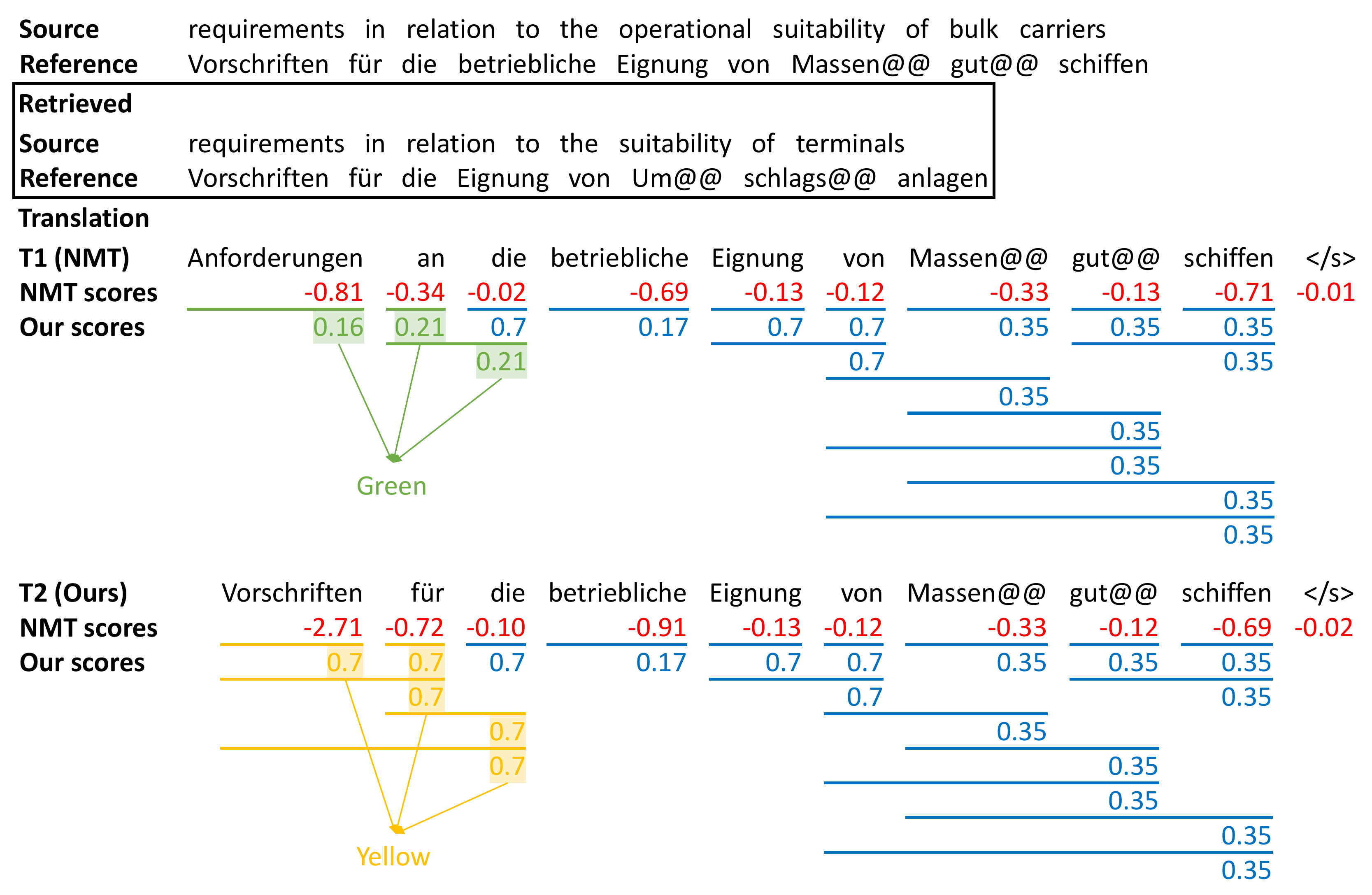}
  \caption{Translation examples. Red scores are log NMT probabilities. Green, yellow and blue scores are scores of matched translation pieces contained only in $T_1$, contained only in $T_2$, contained in both $T_1$ and $T_2$, respectively.}
 \label{t-example}
  \end{figure*}

\begin{table}
\centering

\begin{tabular}{l|rr|rr}
\hline

&\multicolumn{2}{c|}{WMT}&\multicolumn{2}{c}{JRC-Acquis}\\
Similarity&Sent&Percent&Sent&Percent\\
\hline
$[0,0.1)$&0&0\%&4&0.2\%\\
$[0.1,0.2)$&415&13.8\%&141&8.3\%\\
$[0.2,0.3)$&1399&46.5\%&238&14.0\%\\
$[0.3,0.4)$&740&24.6\%&194&11.4\%\\
\hline

$[0.4,0.5)$&281&9.3\%&154&9.1\%\\
$[0.5,0.6)$&113&3.7\%&156&9.2\%\\
$[0.6,0.7)$&29&0.9\%&157&9.2\%\\
$[0.7,0.8)$&10&0.3\%&156&9.2\%\\
\hline

$[0.8,0.9)$&10&0.3\%&252&14.9\%\\
$[0.9,1)$&0&0\%&237&14.0\%\\
$1$&7&0.2\%&0&0\%\\
\hline

\end{tabular}
  \caption{Statistics for similarities between each test sentence and the training set as computed by Equation~\ref{simi1n} for the WMT 2017 en-de task (3004 sentences) and our JRC-Acquis en-de task (1689 sentences).}
  \label{wmt-simi}
\end{table}

 \begin{table}[t]   
 \center
 \begin{tabular}{ll|lll}
 \hline
 &&en-de&en-fr&en-es\\
 
 \hline 
 dev&NMT&44.08&57.26&55.76\\
 &Ours&50.81&62.60&60.51\\
 &1/0 reward&47.70&61.15&58.92\\
 \hline 
 test&NMT&43.76&57.67&55.78\\
 &Ours&50.15&63.27&60.54\\
 &1/0 reward&47.13&62.14&58.66\\
 \hline
 
 \end{tabular}
 \caption{Translation results (BLEU) of 1/0 reward.}
 \label{t-decrease}
 \end{table}

  We also tried our method on WMT 2017 English-to-German News translation task. However, we did not achieve significant improvements over the baseline attentional NMT model, likely because the test set and the training set for the WMT task have a relatively low similarity as shown in Table~\ref{wmt-simi} and hence few useful translation pieces can be retrieved for our method. In contrast, the  JRC-Acquis corpus provides  test sentences that have much higher similarities with  the training set, i.e., much more and longer translation pieces exist.

To demonstrate how the retrieved translation pieces help NMT to generate appropriate outputs, Figure~\ref{t-example} shows an input sentence with reference, the retrieved sentence pair with the highest sentence similarity and outputs  by different systems for this input sentence with detailed scores: 
 log NMT probabilities for each target word in $T_1$ and $T_2$; scores for matched translation pieces  contained in $T_1$ and $T_2$. 
As we can see, NMT assigns higher probabilities to the incorrect translation $T_1$, even though the retrieved sentence pair whose source side is very similar with the input sentence was used for NMT training.

 However, $T_2$ contains more and longer translation pieces with higher scores.
The five translation pieces contained only in $T_2$ are collected from the retrieved sentence pair shown in Figure~\ref{t-example}, which has high sentence similarity with the input sentence. 
The three translation pieces contained only in $T_1$ are also translation pieces collected for the input sentence, but have lower scores, because they are collected from sentence pairs with lower similarities with the input sentence. This shows that computing scores for translation pieces based on sentence similarities is important for the performance of our method. If we assign score $1$ to all translation pieces contained in $G_X$, i.e., use 1/0 reward for translation pieces and non-translation pieces, then the performance of our method decreased significantly as shown in Table~\ref{t-decrease}, but still outperformed the NMT baseline significantly.

\begin{table*} \small 
\centering
\begin{tabular}{ll|llllllll}
\hline
$\gamma$&&0&1&2&5&10&20&50&100\\
\hline
en-de&NMT&5834&3193&1988&1196&717&370&157&75\\
&Ours&5843&5433&3153&1690&933&458&193&86\\
&Ratio (Ours/NMT)&1.00&1.70&1.58&1.41&1.30&1.23&1.22&1.14\\
\hline
en-fr&NMT&6983&3743&2637&1563&812&493&210&118\\
&Ours&7058&5443&3584&1919&968&581&214&134\\
&Ratio (Ours/NMT)&1.01&1.45&1.35&1.22&1.19&1.17&1.01&1.13\\
\hline
en-es&NMT&6500&3430&2292&1346&772&437&182&95\\
&Ours&6516&4589&2970&1652&895&500&196&97\\
&Ratio (Ours/NMT)&1.00&1.33&1.29&1.22&1.15&1.14&1.07&1.02\\
\hline 
\end{tabular}
\caption{$Coun{t_\gamma }$}
\label{t-count}
\end{table*}

\subsection{Infrequent $n$-grams}
The basic idea of our method is rewarding $n$-grams that occur in the training set during NMT decoding.
We found our method is especially useful to help the translation for infrequent $n$-grams.
First, we count how many times a target $n$-gram $u$ occurs in the training set $D_{train}$ as,
\begin{equation}\small 
Occur\left( u \right) = \left| {\left\{ {Y:\left\langle {X,Y} \right\rangle  \in {D_{train}} \wedge u \in uniq\left( Y \right)} \right\}} \right|
\end{equation}
where $uniq\left( Y \right)$ is the set of uniq $n$-grams (up to $4$-grams) contained in $Y$.

Given system outputs $\left\{ {{Z^k}:1 \le k \le K} \right\}$ for the test set $\left\{ {{X^k}:1 \le k \le K} \right\}$ with reference $\left\{ {{Y^k}:1 \le k \le K} \right\}$, we count the number of correctly translated $n$-grams that occur $\gamma$ times in the training set as,
\begin{equation}
Coun{t_\gamma } = \sum\limits_{k = 1}^K {\left| {\psi \left( {\gamma ,{Z^k},{Y^k}} \right)} \right|} 
\end{equation}
where
{\small \begin{align}
&\psi \left( {\gamma ,{Z^k},{Y^k}} \right) = \nonumber \\ &\left\{ {u:u \in \left( {uniq\left( {{Z^k}} \right) \cap uniq\left( {{Y^k}} \right)} \right) \wedge Occur\left( u \right) = \gamma } \right\}
\end{align}}

Table~\ref{t-count} shows $Coun{t_\gamma }$ for different system outputs. As we can see, our method helped little for the translation of $n$-grams that do not occur in the training set, which is reasonable because we only reward $n$-grams that occur in the training set. However, our method helped significantly for the translation of $n$-grams that do occur in the training set but are infrequent (occur less than 5 times). As the frequency of $n$-grams increases, the improvement caused by our method decreased. We analyze that the reason why our method is especially helpful for infrequent $n$-grams is that NMT is trained on the whole training corpus for maximum likelihood and tends to generate more frequent $n$-grams while our method computes scores for the collected translation pieces based on sentence similarities and does not prefer more frequent $n$-grams.

\subsection{Computational Considerations}
\label{comp-plex}

\begin{table}[t]
\center
\begin{tabular}{l|lll}
\hline
&en-de&en-fr&en-es\\
\hline 
Base NMT decoding&0.215&0.224&0.227 \\ \hline
Search engine retrieval&0.016&0.017&0.016\\
TP collection&0.521&0.522&0.520\\
Our NMT decoding&0.306&0.287&0.289\\

\hline
\end{tabular}
\caption{Translation time (seconds).}
\label{t-time}
\end{table}
Our method only collects translation pieces to help NMT for translating a new sentence and does not influence the training process of NMT. Therefore, our method does not    increase the NMT training time. 
Table~\ref{t-time} shows the average time needed for translating  one input sentence in the development set in our experiments.
The search engine retrieval and translation piece  (TP) collection time is computed on a  3.47GHz Intel Xeon  X5690 machine using one CPU. The NMT decoding time is computed using one GPU GeForce GTX 1080.

 \begin{figure}[t]
   \center
   \includegraphics[width=0.4\textwidth]{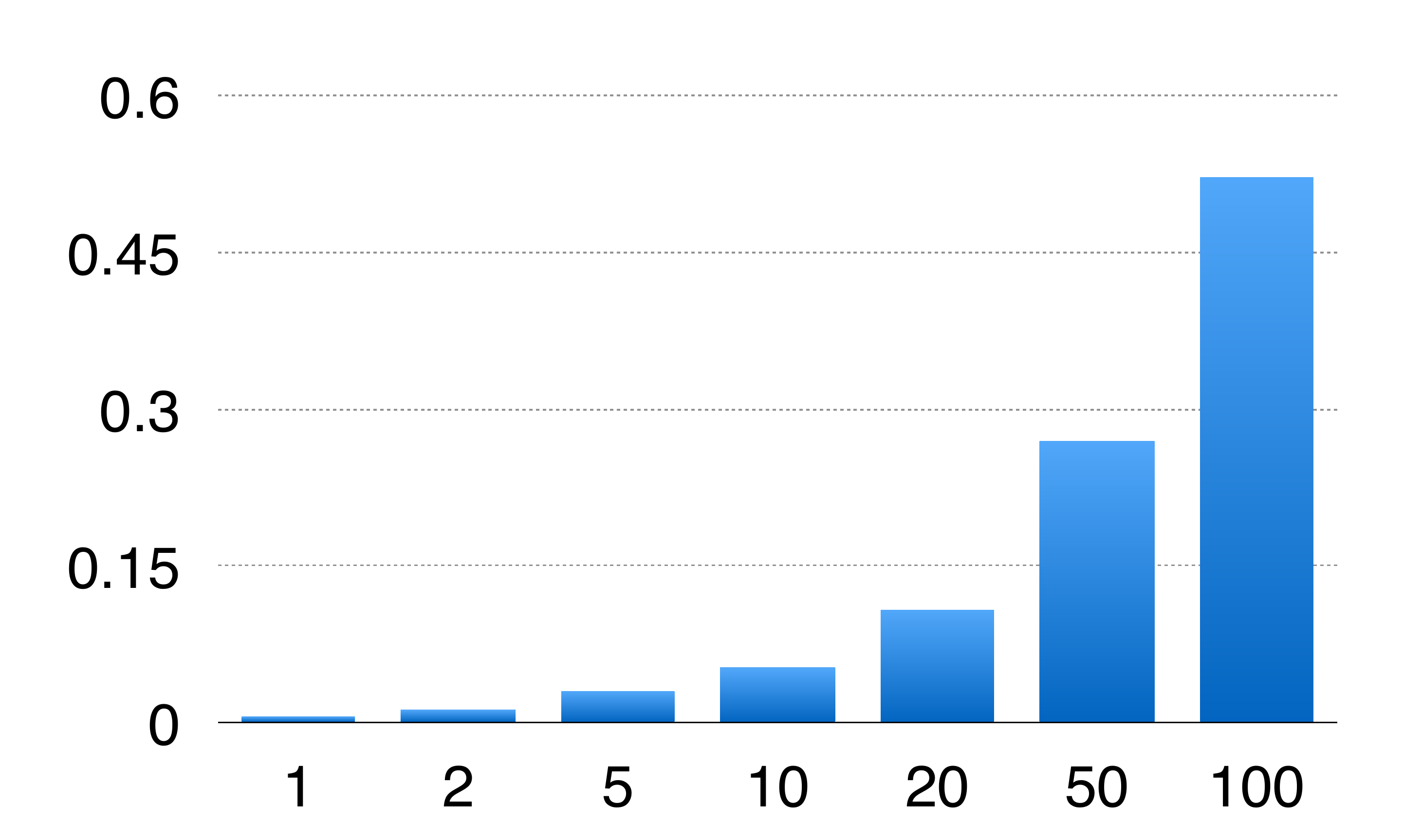}
   \caption{Translation piece collection time (seconds) with different search engine retrieval sizes. }
   \label{f-time}
  \end{figure}
  
 \begin{figure}[t]
   \center
   \includegraphics[width=0.4\textwidth]{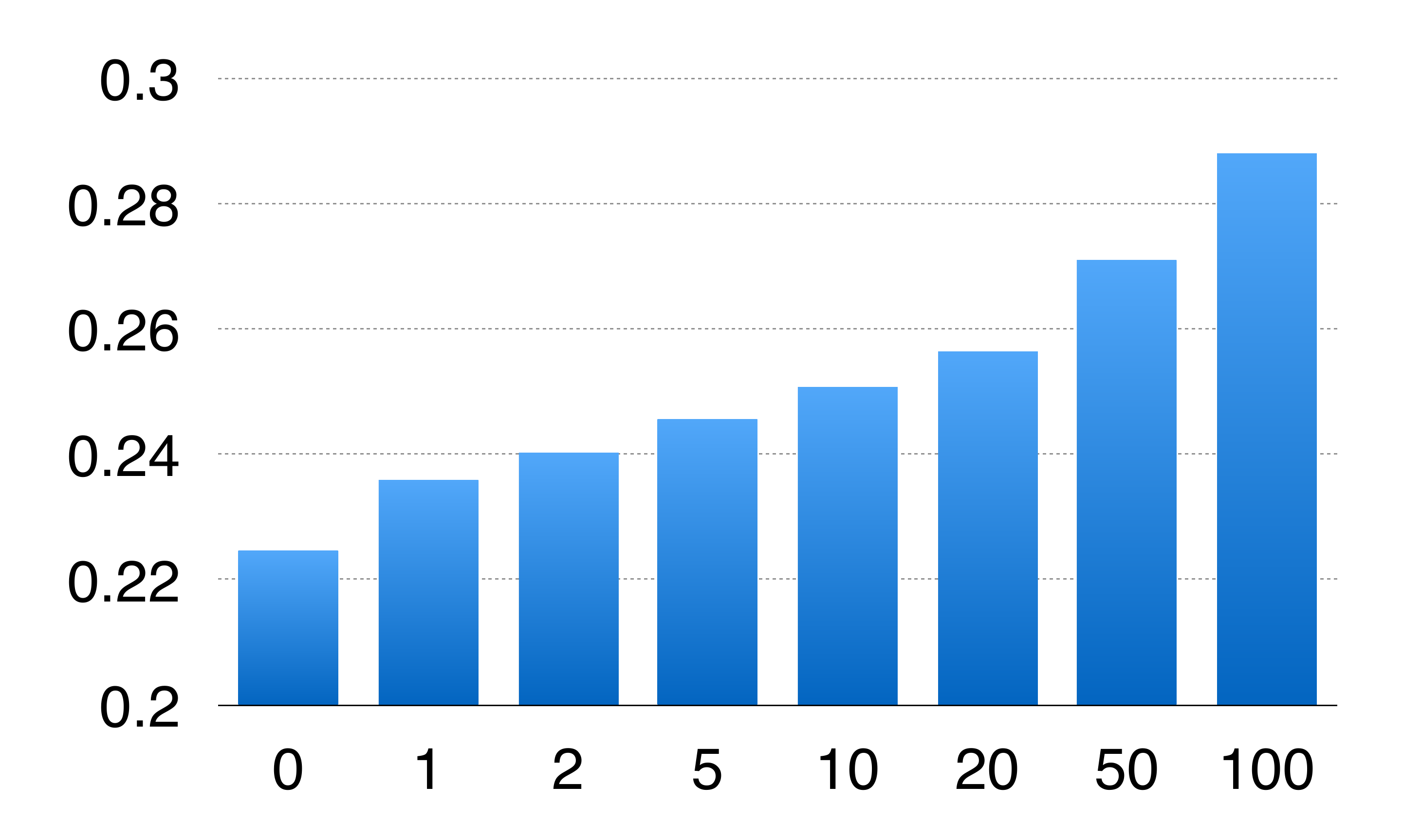}
   \caption{NMT decoding time (seconds) with different search engine retrieval sizes.}
   \label{f-deco}
  \end{figure}
  
   \begin{figure}[t]
     \center
     \includegraphics[width=0.4\textwidth]{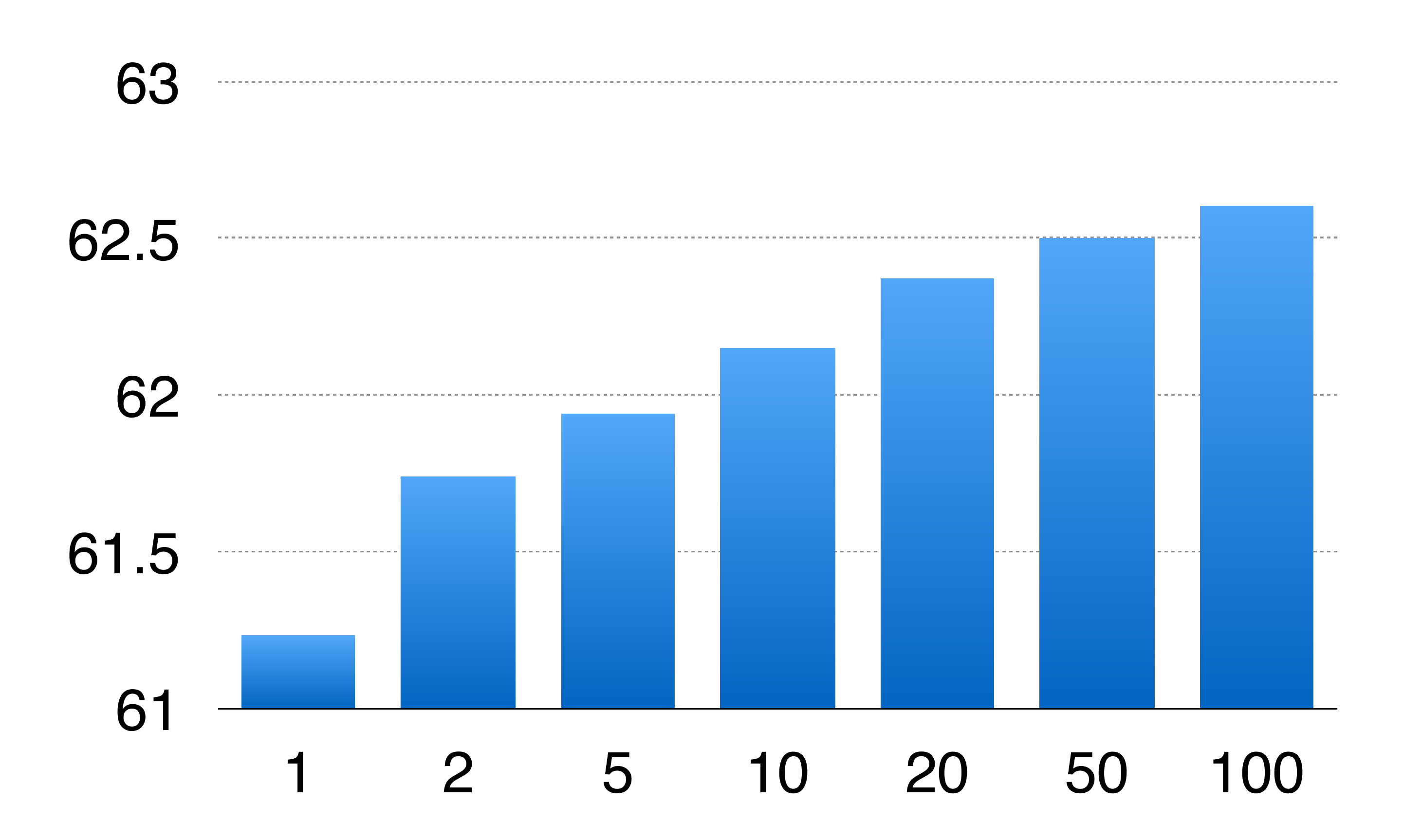}
     \caption{Translation results (BLEU) with different search engine retrieval sizes.}
     \label{f-bleu}
    \end{figure}

As we can see, the search engine retrieval time is negligible and the increase of NMT decoding time caused by our method is also small. However, collecting translation pieces needed considerable time, although our implementation was in Python and could potentially be significantly faster in a more efficient programming language. 
The translation piece collection step mainly consists of two parts: computing the edit distances between the input sentence and the retrieved source sentences using  dynamic programming with time complexity $O\left( {{n^2}} \right)$; collecting translation pieces using Algorithm~\ref{a1} with time complexity $O\left( {4n} \right)$. 

We changed the size of sentence pairs retrieved by the search engine and analyze its influence on translation performance and time. Figure~\ref{f-time}, \ref{f-deco} and \ref{f-bleu} show the translation piece collection time, the NMT decoding time and translation BLEU scores with different search engine retrieval sizes for the en-fr task. 
As we can see, as the number of retrieved sentences decreased, the time needed by translation piece collection decreased significantly, the translation performance decreased much less significantly and the NMT decoding time is further reduced.
In our experiments,
10 is a good setting for the retrieval size, which gave significant BLEU score improvements and caused little increase in the total translation time compared to the NMT baseline.

\subsection{Comparison with SGNMT}
\label{sgnmt}
We compared our method with the search engine guided NMT (SGNMT) model \cite{gu2017search}. We got their preprocessed datasets and tested our method on their datasets, in order to fairly compare our method with their reported BLEU scores.\footnote{Only BLEU scores are reported in their paper.} Table~\ref{t-reported} shows the results of their method and our method with the same settings for the baseline NMT system. As we can see, our method generally outperformed their method on the three translation tasks.

Considering the computational complexity, their method also performs search engine retrieval for each input sentence and computes the edit distance between the input sentence and the retrieved source sentences as our method. In addition, their method runs the NMT model for each retrieved sentence pair to obtain the NMT encoding and decoding information of the retrieved sentences as key-value memory to guide the NMT model for  translating the real input sentence, which changes the  NMT model structure and increases both the training-time and test-time computational cost. Specifically, at test time,  running the NMT model for one retrieved sentence pair  costs the same time as translating the retrieved source sentence with beam size 1. Therefore, as the number of the retrieved sentence pairs increases to the beam size of the baseline NMT model, their method doubles the translation time.

\begin{table}[t]  
\center
\begin{tabular}{ll|lll}
\hline
&&en-de&en-fr&en-es\\
\hline
dev&NMT$_{reported}$&44.94&58.95&50.54\\
&SGNMT$_{reported}$&49.26&64.16&\bf 57.62\\
&NMT&45.18&59.08&50.71\\
&Ours&\bf 50.61&\bf 65.03&57.49\\
\hline
test&NMT$_{reported}$&43.98&59.42&50.48\\
&SGNMT$_{reported}$&48.80&64.60&\bf 57.27\\
&NMT&44.21&59.43&50.61\\
&Ours&\bf 50.36&\bf 65.69&57.11\\
\hline
\end{tabular}
\caption{Comparison with SGNMT.}
\label{t-reported}
\end{table}

\section{Conclusion}
 This paper presents a simple and effective method that retrieves translation pieces to guide NMT for narrow domains. We first exploit a search engine to retrieve sentence pairs whose source sides are similar with the input sentence, from which  we collect and weight translation pieces for the input sentence based on word-level alignments and sentence similarities. Then we use an existing NMT model to translate this input sentence and give an additional bonus  to outputs that contain the collected translation pieces.
 We show our method improved NMT translation results up to 6 BLEU points on three narrow domain translation tasks, caused
 little increase in the translation time,
  and compared favorably to another alternative retrieval-based method with respect to accuracy, speed, and simplicity of implementation.
 \section*{Acknowledgments}
 We thank Jiatao Gu for providing their   preprocessed datasets in Section~\ref{sgnmt}.

\bibliography{naaclhlt2018}
\bibliographystyle{naaclhlt2018}
\end{document}